\pdfoutput=1

\documentclass[11pt]{article}

\usepackage[preprint]{acl}
\usepackage[T1]{fontenc}
\usepackage[utf8]{inputenc}
\usepackage{microtype}

\usepackage{inconsolata}

\usepackage{graphicx}
\usepackage{times}  
\usepackage{algorithm}
\usepackage{algorithmic}

\usepackage{newfloat}
\usepackage{listings}
\DeclareCaptionStyle{ruled}{labelfont=normalfont,labelsep=colon,strut=off} 
\floatstyle{ruled}
\newfloat{listing}{tb}{lst}{}
\floatname{listing}{Listing}
\usepackage{times}
\usepackage{latexsym}

\usepackage[T1]{fontenc}

\usepackage[utf8]{inputenc}

\usepackage{microtype}

\usepackage{inconsolata}

\usepackage{amsmath}
\usepackage{booktabs}
\usepackage{graphicx}
\usepackage{cleveref}
\usepackage{multirow}
\usepackage{pifont}
\usepackage{listings}
\usepackage{caption}
\usepackage{cleveref}
\usepackage[leftcaption]{sidecap}
\usepackage{xcolor}

\title{PuzzleGPT: Emulating Human Puzzle-Solving Ability for Time and Location Prediction}

\author{
Hammad Ayyubi\thanks{Equal Contribution} \quad Xuande Feng\footnotemark[1] \quad Junzhang Liu\footnotemark[1] \quad \\
\textbf{Xudong Lin \quad Zhecan Wang \quad Shih-Fu Chang} \\
Columbia University \\
{\tt\small
   hayyubi@cs.columbia.edu, jl6262@columbia.edu
 }
}
\usepackage{bibentry}

\begin{document}

\maketitle

\begin{abstract}
The task of predicting time and location from images is challenging and requires complex human-like puzzle-solving ability over different clues. In this work, we formalize this ability into core skills and implement them using different modules in an expert pipeline called PuzzleGPT. PuzzleGPT consists of a perceiver to identify visual clues, a reasoner to deduce prediction candidates, a combiner to combinatorially combine information from different clues, a web retriever to get external knowledge if the task can't be solved locally, and a noise filter for robustness. This results in a zero-shot, interpretable, and robust approach that records state-of-the-art performance on two datasets -- TARA and WikiTilo. PuzzleGPT outperforms large VLMs such as BLIP-2, InstructBLIP, LLaVA, and even GPT-4o, as well as automatically generated reasoning pipelines like VisProg\cite{gupta2022visual}, by at least 32\% and 38\%, respectively. It even rivals or surpasses finetuned models.
\end{abstract}

\begin{figure*}[t]
    \centering
    \includegraphics[width=\linewidth]{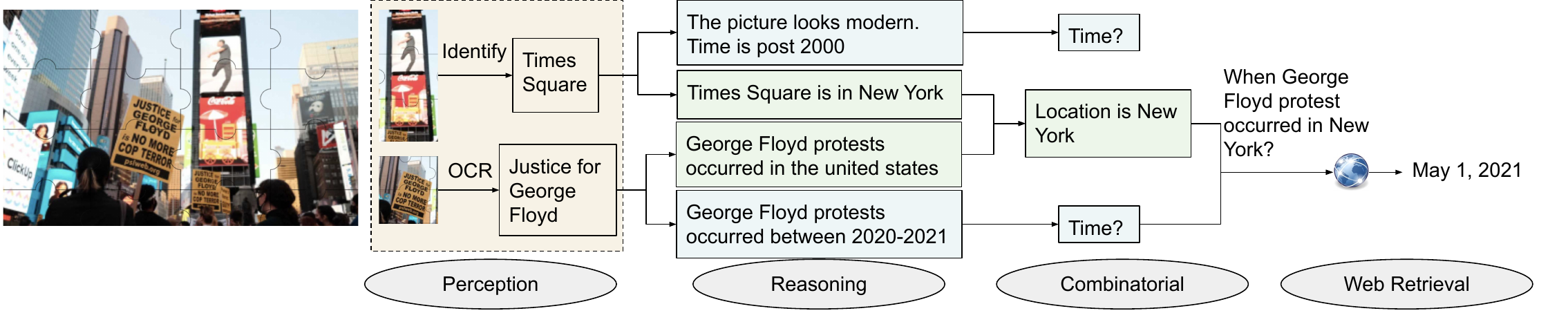}
    \caption{An illustration of different skills (Puzzle-like) required for solving Time and Reasoning prediction tasks.}
    \label{fig:teaser}
\end{figure*}

%
\section{Introduction}

Recent advances in Vision-Language (VL) research have led to models that perform impressively \cite{zhu2023minigpt, li2022mplug, lu2022unifiedio, alayrac2022flamingo, geminiteam2024gemini} on a variety of tasks such as GQA \cite{hudson2019gqa}, VQA v2 \cite{vqa}, VCR \cite{zellers2019vcr}, OK-VQA \cite{okvqa}, Science-QA \cite{lu2022learn}, visual entailment \cite{xie2019visual}. Chain-of-thought reasoning \cite{lu2024mathvista, lu2022learn}. These tasks primarily assess one of, or at most a combination of, perception, reasoning, and outside knowledge retrieval abilities. For example, OK-VQA requires perception and outside knowledge retrieval, and GQA and VCR require perception and commonsense reasoning.

However, humans seamlessly integrate a variety of skills – perception, reasoning, knowledge retrieval, and common sense – to solve complex, multi-step problems.
Tasks and benchmarks that test these diverse skills are crucial for developing models that approach human-level reasoning.
The task of time and place reasoning from images proposed by TARA \cite{fu2022time} takes a step closer to this goal.
It demands a mix of perception, reasoning, combinatorial, and outside knowledge retrieval abilities over multiple steps. It is like solving a puzzle.
For example, in Figure 1, it is required to detect entities such as Times Square and visual text, ``Justice for George Floyd''. Then, a reasoner needs to deduce possible location (New York/United States) and time candidates (post-2000, 2020-2021) from these clues. Next, these candidates need to be combined in various ways to find a candidate at the intersection of all candidates (post 2000 $\cap$ 2020-2021 = 2020-2021). Finally, if the answer is still unclear (2020-2021), a web search is required using the deduced information.

The practical applications of the task stem from its focus on images depicting events that occurred at a specific location and time. This has incredibly useful applications, such as timeline construction, and stitching together news stories from online pictures and social media posts.

Existing works take a direct approach to this nuanced problem. TARA tries to supervise a model directly to predict location and time directly. The hope is that the model will learn to identify time and location discriminative clues implicitly, given appropriate supervisory signals. While the approach might have worked for a limited time and location candidates, the real scenario of hundreds of locations/time with fine differences makes this approach unscalable, and thus impractical. QR-CLIP \cite{shi2023qrclip} additionally tries to incorporate external knowledge in the learning process. However, it oversimplifies the problem and assumes mere retrieval can accomplish the task without relying on specific clues and their combinatorial intersections.

We argue that a complex puzzle-like problem like TARA, requires an equally well thought-out solution. 
To this end, we propose PuzzleGPT. PuzzleGPT abstracts the skills required to solve it into five core abstract ideas: perceiver, reasoner, combiner, noise filter, and knowledge retriever. It represents each with specific modules that perform specific tasks. The perceiver processes visual signals and identifies different entities such as people, buildings, cultural signals, and OCR text. For each of these entities, the reasoner tries to deduce their co-relations with a location and time candidate. 
Integrating clues from multiple entities is crucial for accurate prediction. However, simply combining all clues can introduce noise from irrelevant information, while relying on individual clues might provide insufficient context. To address this challenge, we propose a confidence-based hierarchical combination approach. This approach analyzes clues at increasing levels of granularity: first, individual entities; then, pairs; followed by triplets, and so on, tracking candidate predictions. The process stops once a candidate reaches a threshold vote, efficiently combining entities while minimizing noise.

Apart from being zero-shot, our design choices lend PuzzleGPT desirable properties. Reasoner makes the approach interpretable and thus trustworthy. A hierarchical combination approach makes it not only combinatorial but also noise-resistant. Web retriever infuses the approach with the ability to incorporate world knowledge into the reasoning process. The noise filter adds further robustness.

PuzzleGPT scores state-of-the-art (SOTA) zero-shot performance on TARA, coming close to or surpassing even fine-tuned approaches. We demonstrate that our method outperforms existing SOTA VL models like Instruct BLIP \cite{dai2023instructblip}, BLIP-2 \cite{li2023blip2}, LLaVA \cite{liu2023improvedllava}, by a margin of at least 32\% (standardized location accuracy). It even beats the popular proprietary GPT-4o\cite{OpenAI2024GPT4o}. This highlights current VLMs' inability to simultaneously employ multiple skills to solve a task. We also report superior performance to automatically generated modular pipelines like VisProg, indicating generating an automatic pipeline for this complex task perhaps exceeds their current capabilities. Furthermore, we show that our method generalizes and scores SOTA on another location and time dataset, WikiTilo \cite{zhang2024can}.

We make the following contributions:
\vspace{-4pt}
\begin{itemize}
    \itemsep0em
    \item We propose a novel method, PuzzleGPT, to emulate human puzzle-solving ability for predicting time and location from images.
    \item Our design choices make our approach interpretable, robust, combinatorial, and retrieval augmented.
    \item PuzzleGPT scores SOTA performance on TARA and WikiTilo.
\end{itemize}

\begin{figure*}[t]
    \centering
    \includegraphics[width=\linewidth]{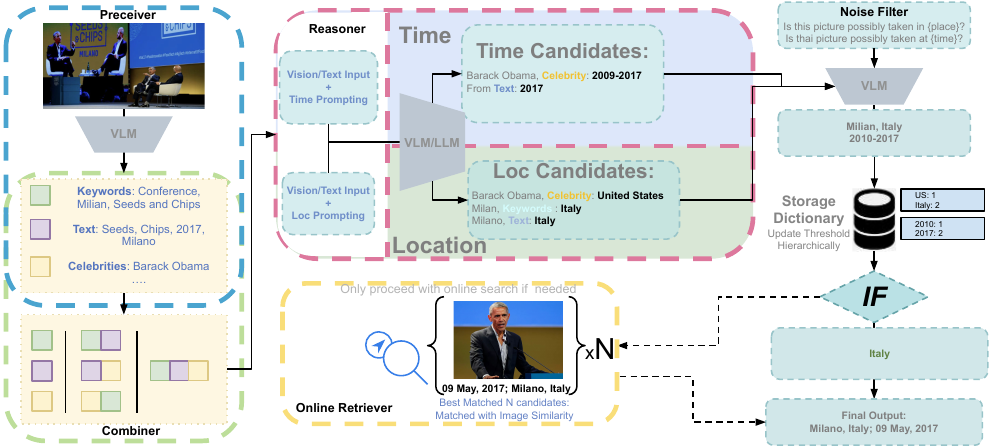}
    \vspace{-0.4cm}
    \caption{PuzzleGPT overview depicting each of the components -- Perceiver, Reasoner, Combiner, Noise Filter, and the Online Retriever. The modular approach makes PuzzleGPT interpretable and robust. All VLMs/LLMs are pretrained and frozen.}
    \label{fig:model}
    \vspace{-0.2cm}
\end{figure*}
\section{Related Work}
\textbf{Vision-Language Models.} Recently, VLMs~\cite{radford2021learning,alayrac2022flamingo,li2023blip2,liu2023visual} have demonstrated remarkable multimodal capabilities through large-scale vision-language training. One family of VLMs such as CLIP~\cite{radford2021learning} typically trains a visual encoder and a text encoder to map visual and text input into a common embedding space. The resulting visual encoders are widely adopted to extract visual features that are fed to LLMs in the other family of work~\cite{alayrac2022flamingo,li2023blip2,liu2023visual,lin2023towards}. For example, LLaVA takes CLIP's visual feature as input and is trained to generate target text. These VLMs with text-generation abilities have achieved superior performance on vision-language datasets.

\textbf{Visual Reasoning Datasets.} Early work like VQA~\cite{vqa} mainly probes perception more than reasoning abilities, while datasets like CLEVR~\cite{johnson2017clevr} focused on compositional reasoning in a controlled synthetic environment. GQA~\cite{hudson2019gqa} pushed towards scene understanding with structured knowledge graphs. Recent work further tackles visual reasoning from different aspects~\cite{zellers2019vcr,han2023infimm,fu2024blink}. However, these datasets either do not require multiple steps of reasoning or lack the depth and breadth of required knowledge. TARA~\cite{fu2022time} and WikiTiLo \cite{zhang2024can}, on the other hand, necessitates multi-step, puzzle-like reasoning over multiple visual clues, combined with external knowledge, posing a unique challenge for existing VLMs. The performance of VLMs such as LLaVA and BLIP-2 remains unsatisfactory on these two datasets.
A recent retrieval-based supervised method~\cite{shi2023qrclip} is proposed to augment CLIP with world knowledge, but it does not yield significant advancement on these tasks either. More importantly, these retrieval-based models' predictions are difficult to interpret.

\textbf{Neural Program Induction / Modular Networks.} Inspired by the need for more composable and interpretable models, research in neural program induction aims to learn programs or modules for solving tasks. Early work explored differentiable neural programmer \cite{neelakantan2015neural}, while Neural Module Networks \cite{andreas2016neural} focused on composing visual modules for reasoning. More recently, VisProg\cite{gupta2022visual} proposed automatic code generation for VQA tasks. However, as our experiments show, automatically generating effective pipelines for intricate problems like TARA remains challenging. PuzzleGPT's expert-designed pipeline, tailored specifically for time and location puzzle-solving, outperforms these automatic approaches, suggesting the importance of domain knowledge and task-specific design for complex reasoning.

\section{Methodology}

We propose PuzzleGPT to emulate human-like puzzle-solving ability. It consists of an expert pipeline consisting of specific modules that represent distinct skills, as illustrated in \Cref{fig:model}. In this section, we describe each of the modules in detail.


\textbf{Perceiver.} 
The perceiver processes visual signals.
Given an image, it will scan the image to find entities of interest, such as celebrities, text, landmarks, or other types of keywords.
In our framework, the Perceiver is a frozen VLM that is prompted with a respective query to extract each entity. For example, text from images is extracted with the query, \textit{`What are the words shown in this image? Short Answer:'}.
By finding the entities, the Perceiver can focus on patches containing specific entities and reason independently. 
This enables it to generate specific textual knowledge about the entities (for a landmark, its name; for text, its Optical Character Recognition; and so on).
We use BLIP-2 as the Perceiver in this work.

\textbf{Reasoner.}  Reasoner in PuzzleGPT is an LLM that deduces time/location candidates from the textual knowledge/clues detected by the Perceiver. For example, given the detected text ``Milano'' in \Cref{fig:model}, the reasoner can deduce Italy as a location candidate. 
The process is demonstrated in \Cref{fig:model}. 
Specifically in this step, we prompt an LLM with a query for each detected entity. As an example, for text in the image, the prompt is \textit{`Does any of the above text contain location information such as city, country, or continent? If yes, please answer that city, country, or continent. Otherwise, answer No. Short Answer:'}.
Considering GPT models' impressive reasoning abilities and cost-effectiveness, we use GPT-3.5 as the Reasoner in this work.

\textbf{Combiner.} As perceiving and reasoning entities independently may provide only a partial view of the correct location/time, there is a need to detect the connections across different entities. For instance, in \Cref{fig:model}, reasoning based on the celebrity name may suggest the location candidate as the United States, even though text clues suggested ``Milano". However, combining all available entities may lead to noisy results. Therefore, we construct a confidence-based combining strategy that strikes a balance between limited information from independent entities and noisy information from all entities. The entity's combination in different ways simulates puzzle-solving human ability.

This combinatorial strategy operates hierarchically: the first level reasons from entities independently, the second combines information from pairs of independent entities, and the third combines all available entities. Each combination generates an answer candidate. The correct candidate, being the most consistent across combinations, is repeatedly predicted while noisy candidates diminish as we go through the hierarchy. To avoid noise becoming dominant in later hierarchies, we stop the iteration once a candidate acquires a threshold number of votes (confidence). Demonstrated in \Cref{hierrarchy} and detailed later in the section.

\textbf{Noise Filter}. In the hierarchical combiner, we enlarged the search space size to find appropriate clue combinations. However, hierarchical combinations will also bring erroneous combinations. Erroneous information will not benefit the reasoning process and can even introduce bias. To address this, we employed a VLM to decide whether the candidate voted by the reasoner is a ``Real Candidate", based on its background knowledge. 
We use BLIP-2 as a Noise Filter as well.

\textbf{Online Retriever.}
VLM/LLMs are, at times, insufficient to reason complicated tasks based only on static knowledge priors obtained through pertaining. 
From a human perspective as well, we try to access online resources once our knowledge is insufficient for a particular task. 
To mimic such an information augmentation for puzzle solving, we allow the model to generate a search query through the Reasoner by providing evidence combination from the combiner. Then, it accesses online dynamic resources through a web search engine. To reduce noise, the online retriever evaluates the relevance between retrievals and the original image through image-to-image/text similarities.
Only retrievals scoring higher than a Retrieval Threshold (RT), are kept.
The retrievals are then fed to the Reasoner to extract the candidate's time/location.
We use CLIP to generate retrieval scores.


\begin{figure}[t]
\captionsetup{skip=3pt}
  \includegraphics[width=\linewidth]{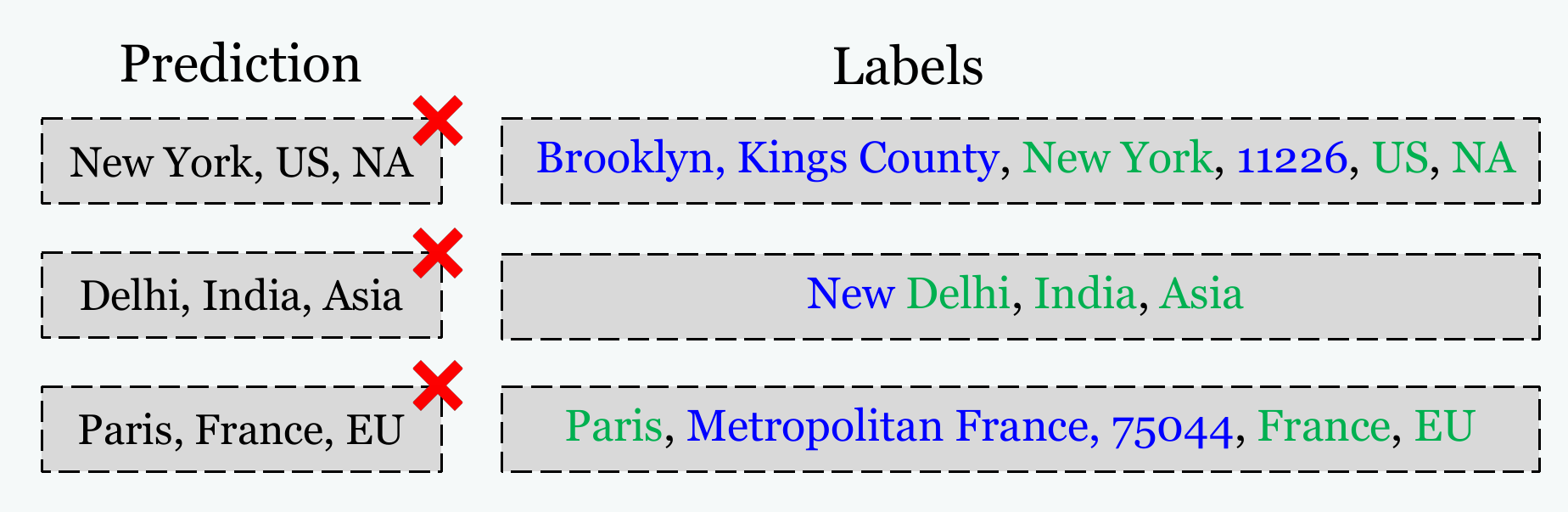}
  \caption {\label{failure} \textcolor{blue}{Extra} location information in the label causes even correct prediction to be classified as \textcolor{red}{wrong} with exact-match accuracy metric. We mitigate this by \textcolor{teal}{label standardization}.}
\end{figure}

\textbf{Confidence-based Hash Thresholding}
As PuzzleGPT's design can generate and obtain significant knowledge and information, it is exposed to a lot of noise. 
They can originate from poor perception, hallucination, or poor web retrievals. 
This needs mitigation.
While Noise Filter aids towards this step, it's not sufficient.

To this end, instead of finding one specific location/time candidate, we instead try to find the location/time candidate with the highest confidence.  
That is, we maintain two hash maps for location and time reasoning, each of which records a candidate accepted by the noise filter. 
To define the state of being 'confident', we set a hyperparameter Hash Threshold, denoted as $HT$. If $HT$ is reached by a candidate, then we know the system would be confident enough that this candidate is the correct answer, and the reasoning process, whichever stage it stands, will (early) stop. If the threshold is never reached, the candidate with the highest confidence will be the output, representing our most confident answer. The hash threshold $HT$ is initially set to $5$.

We detail all prompts and specifics of the Perceiver, Reasoner, Noise Filter, and Online Retriever in \Cref{sec:prompts}.




\section{Experiments}
In this section, we report our results on two datasets: TARA and WikiTilo. We have run all the experiments on a single Nvidia RTX 4090 24 GB GPU.


\subsection{TARA} 
TARA is sourced from the New York Times and requires time and location prediction for images depicting newsworthy events. In total, there are around 1.5K samples in the test and validation set. The label set is open-ended with a unique label for each sample. 

\subsubsection{Metric}
The open-ended nature of labels in TARA makes evaluation challenging. Two metrics were proposed originally -- Accuracy and Example-F1. 
Accuracy measures the exact match of the prediction with the label. While this works for time evaluation where the labels are properly formatted (YYYY-MM-DD), location evaluation leads to unreliable results as the labels are highly unstructured. As illustrated in Fig \Cref{failure} in addition to city, country, and country, some labels contain additional information such as Pin Code, county name (Kings County), and geographical area name (Metropolitan France). This causes even correct predictions to be incorrectly classified as wrong. To address this, we standardize all locations into city, country, and continent using GeoPy \footnote{\url{https://geopy.readthedocs.io/en/stable/}}. Further, if the label contains a specific area within a city (e.g. Times Square or Central Park), we keep that to not lose location precision. We use these formatted labels for measuring accuracy and call it Standardized Accuracy (Std. Acc). 

To measure partial correctness -- only correct year or only correct continent and country -- TARA proposes Example-F1 metric. It is defined as follows:
\begin{equation}
\qquad\qquad\ \ \ \ 
\text{ Example-F1 } =  
\frac{2|GT\cap Pred|}{|GT|+|Pred|} \nonumber
\qquad\qquad\qquad
\end{equation}

As the score is inversely dependent on $|Pred|$, shorter predictions are unduly rewarded. For example, a model that predicts only year scores abnormally high Example-F1. We mitigate this bias by adding a brevity penalty, following NLP literature \cite{bleu_2002}:

\begin{equation}\ 
\text{Example-F1}^\beta =
e^{-(\frac{|Pred|}{|GT|}-1)^+} \cdot \text{Example-F1} \nonumber
\ \ \ \ \ \ \ 
\end{equation} 
We use X-F1 for Example-F1 from here onwards. All our experiments and ablations on TARA use the more correct X-F1$^\beta$ and Location Std. Acc metric, except for \Cref{tab:tara_main_results_finetune} where we use X-F1 and Location Acc for direct comparison against previously reported results. 

\begin{table}[t]
  \centering
  \captionsetup{skip=3pt}
  \setlength{\tabcolsep}{4pt}
  \fontsize{9}{11}\selectfont
  \begin{tabular}{l|cccc}
    \toprule
         &    \multicolumn{2}{c}{\textbf{Time}}&\multicolumn{2}{c}{\textbf{Location}}\\
    \textbf{Model}  &    \textbf{ Acc(\%)}    & \textbf{X-F1$^\beta$} &\textbf{Std. Acc(\%)} & \textbf{X-F1$^\beta$}  \\
    \midrule
    BLIP2        &  0.30   & 32.27                & \textit{17.41} & 43.59                 \\
    LLaVA        &  0.23   & \textit{43.26}       & 7.85           & 25.92                \\
    GPT4o        &  0.30   & 21.94                & 16.62          & \textit{47.16}        \\
    InstructBLIP &  0.00    & 33.83                & 16.69          & 26.05                    \\
    \midrule
    IdealGPT     &  0.27    & 26.83                &  9.95          & 25.70                    \\
    VisProg      &  0.00    & 18.52                & 0.00           & 4.74                  \\
    ViperGPT      &  0.00    & 36.05                & 1.67          & 34.65                  \\
    \midrule
    PuzzleGPT         &  \textbf{0.30}   & \textbf{43.72}    & \textbf{22.99}  & \textbf{51.04} \\
    \bottomrule
  \end{tabular}
  \caption{\label{tab:tara_main_results_zeroshot}
   We compare PuzzleGPT 
   to SOTA zero-shot generative VLMs on TARA. PuzzleGPT outperforms all prior methods, scoring SOTA performance.
  }
\end{table}

\begin{table}[t]
  \centering
  \captionsetup{skip=3pt}
\fontsize{9}{11}\selectfont
\setlength{\tabcolsep}{4pt}
  \begin{tabular}{l|cccc}
    \toprule
         &       \multicolumn{2}{c}{\textbf{Time}}&\multicolumn{2}{c}{\textbf{Location}}\\
    \textbf{Model}  &\textbf{Acc(\%)}           & \textbf{X-F1}    & \textbf{Acc(\%)} & \textbf{X-F1} \\
    \midrule
    CLIP       & 0.46   &  39.90        & 11.11             & 44.96     \\
    CLIP+      & 1.00   &  43.09        & 15.72             & 49.74     \\
    CLIP+Seg   & 0.92   &  42.82        & 16.46             & 50.52     \\
    QR-CLIP    & \textbf{3.53}   &  \textbf{47.89}        & 19.31             & 50.96     \\
    \midrule
    PuzzleGPT  &0.30  &\textit{43.72}  & \textbf{22.99$^*$ }  &\textbf{56.11}   \\
    \bottomrule
  \end{tabular}
  \caption{\label{tab:tara_main_results_finetune}
   PuzzleGPT comparison against representative classification models reported in prior works. All are finetuned except CLIP. * denotes Std. Acc. PuzzleGPT outperforms finetuned methods on location reasoning while recording comparable performance on time prediction.}
   \vspace{-0.3cm}
\end{table}

\subsubsection{Baselines}
In addition to comparing PuzzleGPT against previously reported approaches on TARA, we also evaluate it against recent VLMs to provide a comprehensive comparison and valuable insights.

\noindent\textit{\textbf{Large Vision Language Models.}}
We evaluate the following VLMs: BLIP-2, InstructBLIP, LLaVA, and GPT-4o. These models leverage the power of LLMs for visual reasoning, thereby acquiring extensive knowledge and reasoning abilities. They represent single-stop solutions for complex tasks.

\noindent\textbf{\textit{Code Based Modular Approaches.}}
We also compare PuzzleGPT to methods that generate modular code for various VL tasks, such as VisProg. These methods serve as references for automatic pipelines, contrasting with our expert pipeline. Additionally, we compare against IdealGPT, which aims to enhance robustness in automatic pipelines through an iterative pipeline.

\begin{table}[t]
  \centering
\captionsetup{skip=3pt}
  \setlength{\tabcolsep}{4pt}
  \fontsize{9}{11}\selectfont
  \begin{tabular}{lcc}
    \toprule
    \textbf{Ablations}  & \textbf{Time X-F1}$^\beta$& \textbf{Location X-F1}$^\beta$\\
    \midrule
    PuzzleGPT  &\textbf{43.72} &\textbf{51.04}\\
    \hspace{3pt} 1st Hier Only      &  42.62 & 46.68  \\
    \hspace{3pt} 3rd Hier Only      &  42.72  & 45.21 \\
    \bottomrule
  \end{tabular}
  \caption{\label{tab:ablation_hier}
    Hier: Hierarchy. Confidence-based hierarchical combination is critical. PuzzleGPT outperforms simpler methods by avoiding incomplete information from 1st Hier Only and noise from 3rd Hier Only.
  }
\end{table}
\begin{figure}[t]
\centering
\captionsetup{skip=3pt}
\scalebox{0.5}{
\includegraphics{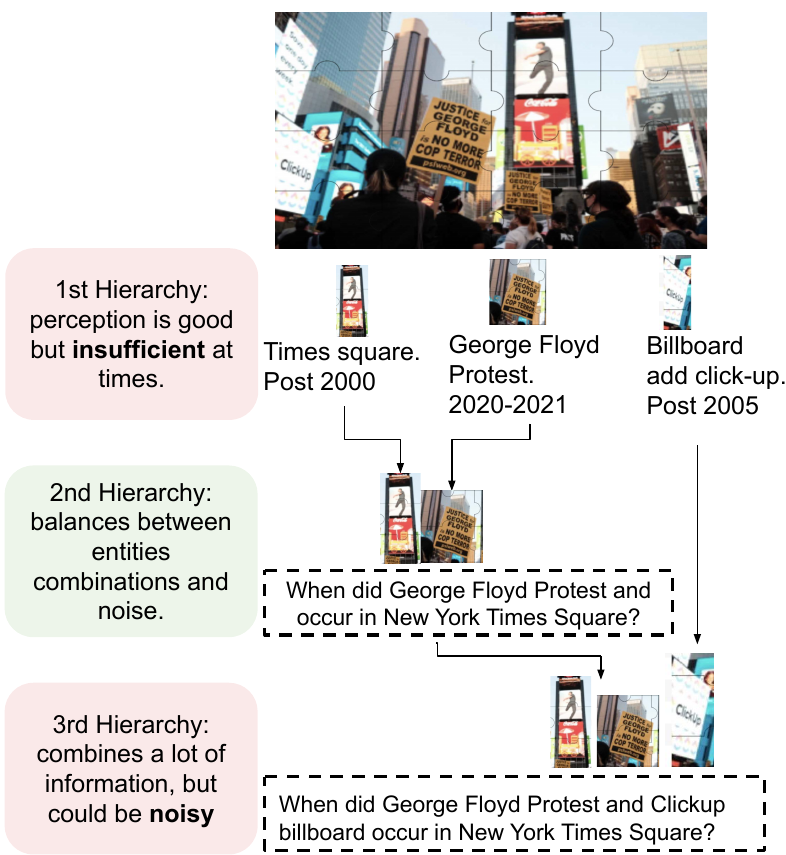}}
  \caption {\label{hierrarchy} An illustration of the importance of confidence-based hierarchical combination of information. 1st Hierarchy leads to a scarcity of information, while 3rd Hierarchy is noisy, and underscores the need for a confidence-based hierarchical combination.}
\end{figure}

\subsubsection{Results}




We compare PuzzleGPT against zero-shot VLMs in \Cref{tab:tara_main_results_zeroshot} and finetuned approaches in \Cref{tab:tara_main_results_finetune}. We make the following observations:

\noindent\textit{\textbf{PuzzleGPT records state-of-the-art performance.}} PuzzleGPT outperforms all methods, including the popular GPT-4o model, for both location and time prediction. It's especially skilled at location prediction: $>$30\% Std. Acc. improvement over next best method (BLIP-2).

\noindent\textit{\textbf{PuzzleGPT is more skilled than single-stop Large VLMs.}} PuzzleGPT's strong improvements over all VLMs indicate their limitation in leveraging diverse skills to accomplish this complex task.

\noindent\textit{\textbf{PuzzleGPT's expert pipeline is better at puzzle-like tasks than automatic pipelines.}} From Visprog's inferior performance, we conclude that automatic pipelines are 1) constrained by the types of skills they can apply and 2) the search space for the optimum pipeline in puzzle-like tasks is so large that they generate suboptimal code.

\noindent\textbf{\textit{PuzzleGPT comes close to or surpasses finetuned performance.}} PuzzleGPT's effectiveness and strong performance are highlighted by the fact that it achieves $>$10\% X-F1 improvement over the best fine-tuned approach (QR-CLIP).

\subsubsection{Ablation Studies}

\begin{figure}[t]
\captionsetup{skip=-10pt}
  \includegraphics[width=0.48\linewidth]{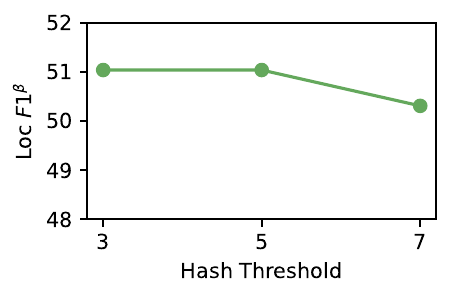} 
  \includegraphics[width=0.5\linewidth]{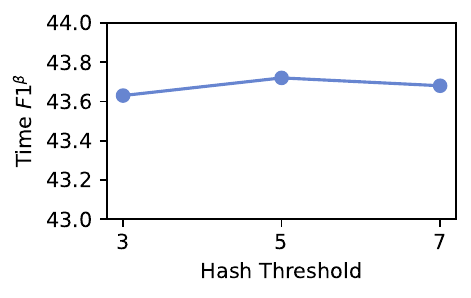} \\
  \includegraphics[width=0.48\linewidth]{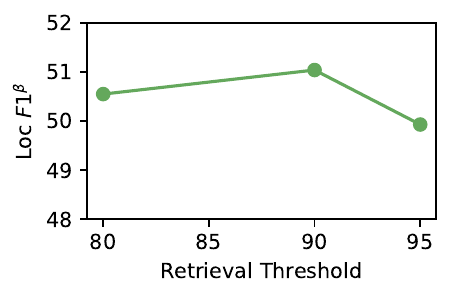} 
  \includegraphics[width=0.5\linewidth]{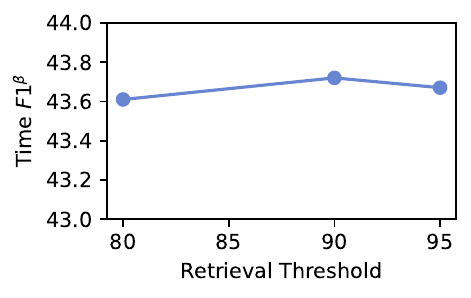} \\
  \caption {\label{RTHT} Top: Ablation on Hash Threshold (HT): peak performance at 5, with noisy performance on both lower or higher HT. Bottom: Ablation on Retrieval Threshold (RT): retrieval is best at 90, with either side of it leading to noisy retrieval. Loc F1$^\beta$ is Location X-F1$^\beta$. Time F1$^\beta$ is Time X-F1$^\beta$.}
\end{figure}

\begin{figure*}[t]
\centering
\captionsetup{skip=3pt}
  \includegraphics[width=0.48\linewidth]{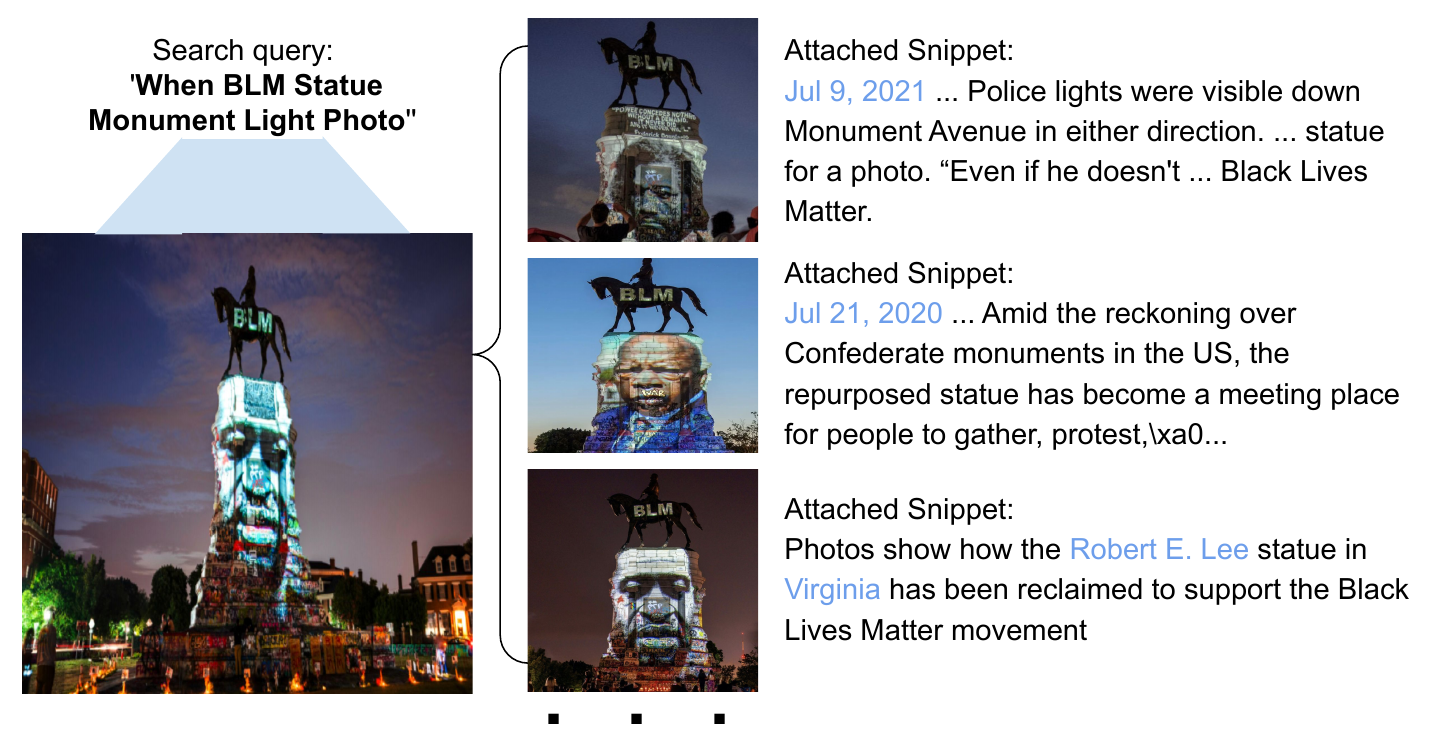}
    \includegraphics[width=0.48\linewidth]{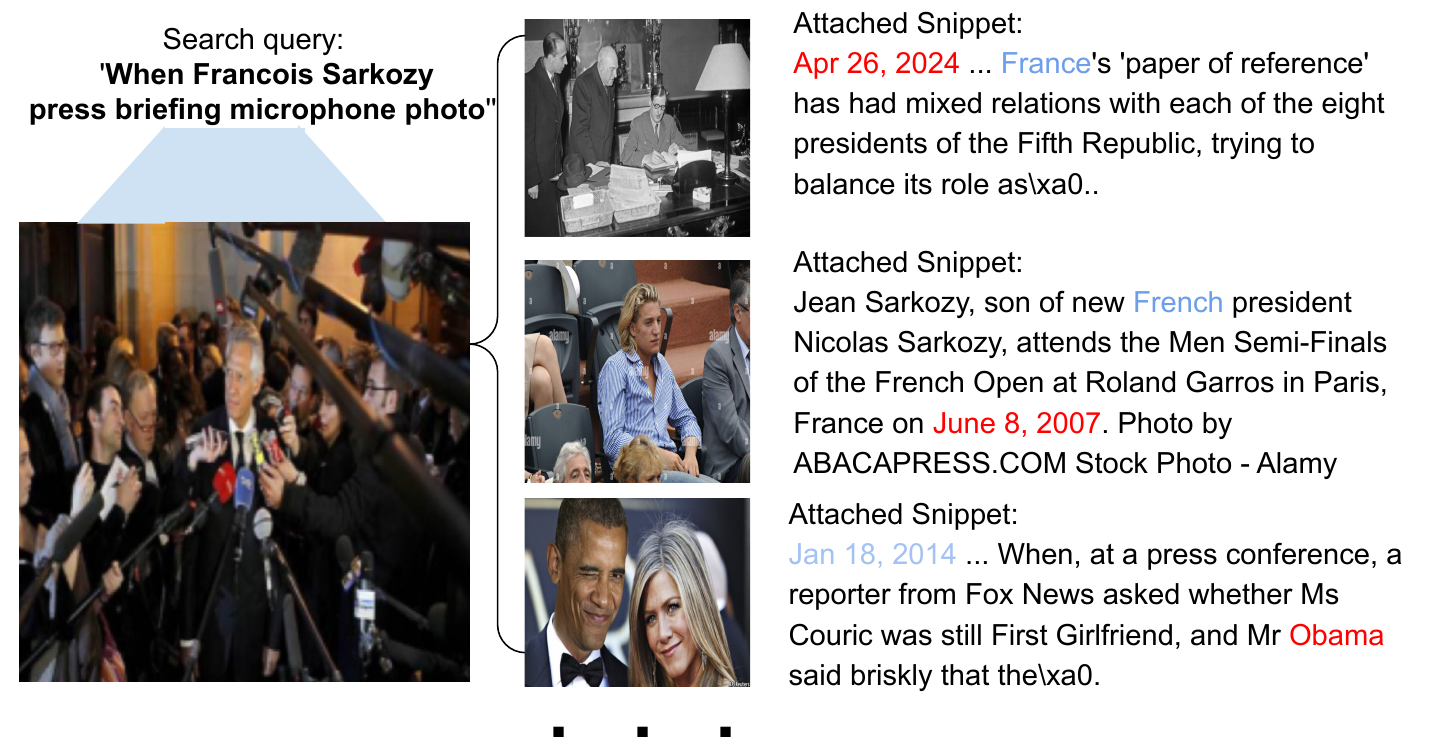}
  \caption {With specific and clear clues, our model can retrieve high-quality web content while generic images tend to retrieve noisy content.}
  \label{fig:retr_pos}
\end{figure*}
\begin{table}[t]
  \centering
    \captionsetup{skip=3pt}
  \setlength{\tabcolsep}{4pt}
  \fontsize{9}{11}\selectfont
  \begin{tabular}{lcc}
    \toprule
    \textbf{Ablations}  & \textbf{Time X-F1}$^\beta$& \textbf{Location X-F1}$^\beta$ \\
    \midrule
    PuzzleGPT   & \textbf{43.72}  & \textbf{51.04}\\
    \hspace{3pt} - w/o Filtering &  39.27 & 48.77 \\
    \hspace{3pt} - w/o Retrieval     &  42.63  & 43.30\\
    \bottomrule
  \end{tabular}
  \caption{\label{tab:tara_ablation_filter_web}
    Noise Filter and Retriever ablation. Performance drop if we remove either of them, underscoring their importance to PuzzleGPT.
  }
\end{table}

\begin{table}[t]
  \centering
    \captionsetup{skip=3pt}
  \setlength{\tabcolsep}{4pt}
  \fontsize{9}{11}\selectfont
  \begin{tabular}{lcc}
    \toprule
    \textbf{Ablations}  & \textbf{Time X-F1}$^\beta$& \textbf{Location X-F1}$^\beta$\\
    \midrule
    PuzzleGPT (I-I Retrieval)  & \textbf{43.72}  & \textbf{51.04}\\
    \hspace{3pt} - I-T Retrieval & 43.47 &   50.95   \\
    \bottomrule
  \end{tabular}
  \caption{\label{tab:tara_ablation_it_matching}
    I: Image. T: Text. Retrieval is best served by image-image matching. Replacing it with image-text retrieval reduces performance.}
   \vspace{-0.3cm}
\end{table}

\begin{figure}[t]
\centering
\captionsetup{skip=-3pt}
  \includegraphics[width=0.8\linewidth]{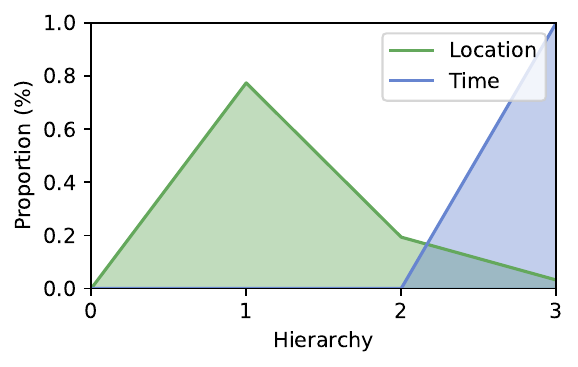}
  \caption {Distribution of endpoints of hierarchy. Time prediction is more complex than location prediction, with most location queries resolved in the first hierarchy while most time queries reach the third hierarchy.}
  \label{fig:hierarchy_dist}
\end{figure}
\begin{figure}[t]
\centering
\captionsetup{skip=-3pt}
  \includegraphics[width=0.8\linewidth]{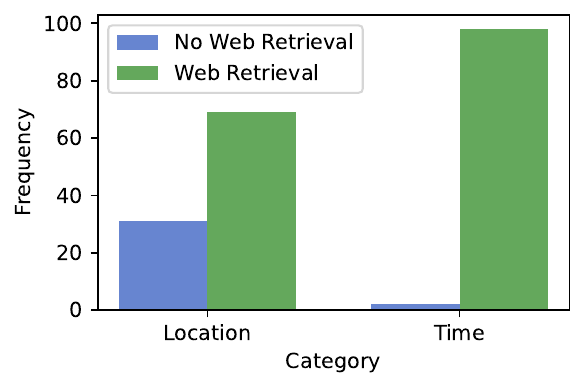}
  \caption {
  Frequency of retrieval and no retrieval between time and location queries. Almost all time queries require web retrieval, highlighting the complexity of time prediction compared to location prediction.
  }
  \label{fig:retrieval_vs_no_retrieval}
\end{figure}

We investigate PuzzleGPT from different axes to thoroughly analyze its modules.
 The choice of ablations metrics is explained in \Cref{sec:metrics}.

\noindent\textit{\textbf{Confidence-based hierarchical combination is crucial.}} To understand the importance of hierarchical combination, we compare our approach in \Cref{tab:ablation_hier} to simple ablations that 1) do not combine information from entities (1st Hier Only), and 2) combine information from all entities in one go (3rd Hier Only). PuzzleGPT outperforms both. \Cref{hierrarchy} illustrates the underlying reason: 1st Hier only results in incomplete information and 3rd Hier is noisy. These results demonstrate that a confidence-based hierarchical combination is crucial to carve a middle path between incorporating signals from different puzzle pieces and reducing noise.

\noindent\textit{\textbf{Confidence thresholding matters in hierarchical combination.}} \Cref{RTHT} shows that the best performance is reached at threshold=90, with inferior scores for both lower and higher thresholds. This implies low confidence threshold allows noisy candidates to be predicted, while a higher threshold results in more pipeline iterations, thereby introducing additional noisy candidates. 

\noindent\textit{\textbf{Web retrieval augments PuzzleGPT with external knowledge.}} As reported in \Cref{tab:tara_ablation_filter_web}, not retrieving external knowledge from the internet degrades performance by 1.09\% in Time X-F1$^\beta$ and 7.74\% in Location X-F1$^\beta$. \Cref{fig:retr_pos} further illustrates the importance of retrieval, especially for time prediction.

\noindent\textit{\textbf{Retrieval is sensitive to thresholding.}} \Cref{RTHT} plots the model performance against different values of retrieval threshold. The performance peaks at 90, implying the lower threshold is noisy and the higher threshold leads to information bottleneck.

\noindent\textit{\textbf{Retrieval is best served by image-image matching.}} \Cref{tab:tara_ablation_it_matching} reports the performance drop by replacing image-image retrieval with image-text retrieval. Specifically, we observe a drop of 0.25\% in Time X-F1$^\beta$, and 0.09\% in Location X-F1$^\beta$, indicating that it's a suboptimal strategy for this task.

\noindent\textit{\textbf{Time prediction is more complex than location prediction.}} We observe from \Cref{fig:hierarchy_dist} that the majority of queries for location finish in the first hierarchy, while almost all queries for time reach the third hierarchy. This demonstrates that location prediction is doable from individual visual clues, while time prediction requires more combinations of clues to arrive at a candidate. Further, \Cref{fig:retrieval_vs_no_retrieval} reveals that almost all queries for time need web retrieval. All this points to a higher complexity of time prediction.

\noindent\textit{\textbf{Noise filtering is critical.}} From 
\Cref{tab:tara_ablation_filter_web}, we observe that eliminating the noise filtering module leads to a performance drop of 4.45\% in Time X-F1$^\beta$ and 2.27\% in Location X-F1$^\beta$. This highlights the importance of noise filtering.

We further compare performance against an open-sourced Reasoner in \Cref{sec:llama_result} and explore BLIP-2 alternatives for Perceiver in \Cref{sec:perceiver_ablation}.

\subsubsection{Qualitative Analysis}

\begin{figure*}[t]
  \centering
  \captionsetup{skip=-2pt}
  \includegraphics[width=0.8\linewidth]{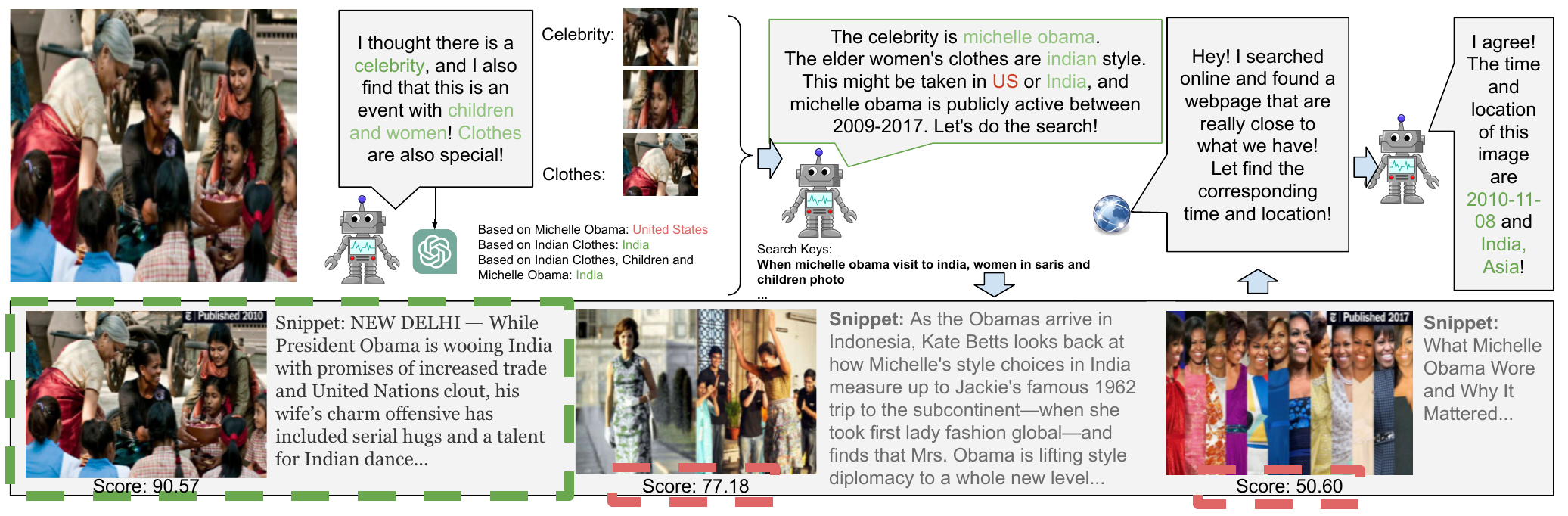} 
  \caption {\label{pnsamples} We illustrate a positive case study for PuzzleGPT depicting the reasoning process it follows to arrive at the prediction. More positive and negative samples are available in the supplementary section.}
\end{figure*}

\begin{table*}[t]
  \centering
    \captionsetup{skip=3pt}
  \setlength{\tabcolsep}{4pt}
  \fontsize{9}{11}\selectfont
  \begin{tabular}{lccccccccc}
    \toprule
         &   \multicolumn{3}{c}{\textbf{Time}}&\multicolumn{3}{c}{\textbf{Country}}&\multicolumn{3}{c}{\textbf{Region}} \\
    \midrule
    \textbf{Models}  &\textbf{Acc(\%)}     & \textbf{Prec}     & \textbf{F1}     & \textbf{Acc(\%)} & \textbf{Prec}     & \textbf{F1} & \textbf{Acc(\%)} & \textbf{Prec}&\textbf{F1}\\
    \midrule
    OpenFlamingo Test   & 27.70   &  26.36 & 11.49   & 3.89  & 3.69   & 2.18 
                              & 4.72    &  8.62  & 4.72  \\
    OpenFlamingo-VQA          & 31.59   &  30.36 & 28.60   & \textbf{48.88} & \textit{53.78} &\textit{ 41.19}                           &22.49 & 30.49 & 18.64\\
    OpenFlamingo-VQA CoT      & 35.21   &  29.36 & 28.42   & 40.3  & 45.24 & 33.17                           & \textit{24.04} & \textit{39.39} & \textit{19.27}\\
    LLaMA-AdapterV2-Instr$^a$   &\textit{ 58.02}     &  28.04 & 32.88   & 23.05 & 52.64 & 18.66                           & 19.07 & 26.59 & 13.01\\
    LLaMA-AdapterV2-Instr$^b$   & 34.34     &  \textit{58.59} & \textit{37.70}   & \textit{45.62} & 51.57 & 35.50                           & 11.12 & 10.05 & 5.99\\
    
    ViperGPT  & 29.52   &  35.27 & 21.50   & 20.32  & 21.96    & 14.98   & 26.03 & 25.88                           & 21.54\\
    
    IdealGPT  & 32.16   &  29.20 & 26.30   & 4.32  &   4.10  &  2.21  & 9.13 & 9.85& 7.68\\
    Frequency baseline  & 25.07   &  25.29 & 23.27   & 3.33  & 2.95    & 2.88   & 12.53 & 12.59                           & 12.25\\
    \midrule
    PuzzleGPT(Ours)             & \textbf{71.90}   &  \textbf{70.63}     & \textbf{72.61}       & 43.65  & \textbf{72.78}  & \textbf{49.79} & \textbf{62.06} &    \textbf{79.22}         & \textbf{68.18}\\
    \bottomrule
  \end{tabular}
  \caption{\label{tab:wikitilo}
PuzzleGPT generalizes to WikiTilo dataset, scoring state-of-the-art performance in almost all the metrics.
  }
\end{table*}

We conducted case studies on TARA in \Cref{pnsamples} 
to qualitatively analyze PuzzleGPT's effectiveness. 
The example demonstrates the interpretability of our approach, modularly depicting the reasoning process and grounding each inference in entities or their combinations. For example, location inference in India is deduced from cultural clues (dress) in the image.
We also noticed that in cases where the images contain less clear/informative clues, our model can fail to discover and ground clues.
In particular, we observe that failure typically occurs when the given image lacks unique landmarks, events, or people.
This is partly illustrated in \Cref{fig:retr_pos} where lack of specificity leads to generic search queries and noisy retrievals.
We demonstrate this further with examples in \Cref{sec:qualitative_analysis}.

\subsection{WikiTilo}


To demonstrate the generalization of our approach, we also report PuzzleGPT's performance on another location/time reasoning dataset, WikiTilo. It contains \textasciitilde 600 images in the test set with a focus on identifying sociocultural cues to predict time/location. While TARA evaluates predictions on open-ended labels, WikiTilo's labels are multi-choice. For location, the evaluation is divided into Country, with 30 multiple choice labels, and Region, with 8 distinct labels. For Time, the labels are divided into four time periods. Since the labels are multi-choice, the predictions are simply scored on accuracy, precision, and F1 score. The absence of a partial correctness case alleviates the need for the X-F1 metric. We compare PuzzleGPT against previously reported models on WikiTilo in \Cref{tab:wikitilo}. We run additional baselines on WikiTilo and report the results in \Cref{sec:additional_wikitilo_baselines}.

From \Cref{tab:wikitilo}, we score state-of-the-art performance on WikiTilo for time and region prediction. 
Specifically, our method improves time Acc. and F1. by $+23.9\%$ and $+123.5\%$ respectively, over the next best method. Region Acc. and F1. are boosted by $+158.2\%$ and $+101.1\%$ respectively. 
For country prediction, our Acc. is slightly worse ($-10.7\%$), but we still outperform the previous best F1 by $+68.3\%$. Our lower Acc can be explained by the inconsistencies in the baseline numbers. 
They score lower on the Regions category with 8 choices but higher on the Countries category with 30 choices, which contradicts the expected pattern of fewer choices leading to better performance. Besides, if a model can predict the correct country (e.g., USA), it should also predict the correct region (e.g., North America), but the scores don't align with this logic.


Indeed, PuzzleGPT predicts time significantly better on WikiTilo than on TARA, 
highlighting the challenge of open-ended time prediction in TARA.

\section{Conclusion}
This work proposes a modular and iterative puzzle-solving method -- PuzzleGPT -- for predicting time and location from images. 
It consistently outperforms current SOTA VLMs on TARA and WikiTilo. The extensive experiments and ablations demonstrate and analyze its capabilities. 
Further, we plan to release the code upon acceptance in an effort to increase community engagement.

\section*{Limitations}
While PuzzleGPT demonstrates strong performance on time and location prediction tasks like TARA and WikiTilo, it's important to acknowledge its limitations. The model's architecture is specifically tailored for puzzle-like reasoning scenarios, and its performance on tasks with different structures or knowledge requirements remains unexplored. Furthermore, the current reliance on GPT for reasoning introduces dependencies on proprietary models, potentially limiting accessibility and introducing inherent biases. Future work will explore alternative reasoning modules and evaluate PuzzleGPT's generalization ability across diverse visual reasoning tasks.

\bibliography{custom}

\clearpage
\appendix
We provide additional details in this section which further elucidate the claims and contributions of the paper. It's divided into the following sections:

\begin{itemize}
    \item Prompts Details. (Appendix \ref{sec:prompts})
    \item Open-Source Llama as the Reasoner (Appendix \ref{sec:llama_result})
    \item Perceiver Ablation (Appendix \ref{sec:perceiver_ablation}) 
    \item Experiments on a Small Clean Subset (Appendix \ref{sec:experiment_small})
    \item Qualitative analysis (Appendix \ref{sec:qualitative_analysis})
    \item Additional WikiTilo Baselines (Appendix \ref{sec:additional_wikitilo_baselines})
    \item Ablation Metrics (Appendix \ref{sec:metrics})\\
\end{itemize}

\section{Prompts Details}
\label{sec:prompts}
We list the most prominent prompts used by us in our experiments.
\subsection{Perceiver Prompts}

\definecolor{backcolor}{rgb}{0.95,0.95,0.92}




\noindent\textit{Keywords}
 \begin{lstlisting}[backgroundcolor = \color{backcolor}, basicstyle=\small, breaklines=true, numbers=none]
"""
USER: <image>
What are the keywords of this image? Answer with the keywords only and separate with commas. 
Short Answer: ASSISTANT: 
"""
\end{lstlisting}

\noindent\textit{Text OCR}
 \begin{lstlisting}[backgroundcolor = \color{backcolor}, basicstyle=\small, breaklines=true, numbers=none]
"""
USER: <image>
What
are the words shown in this image. 
Short Answer: ASSISTANT: 
"""
\end{lstlisting}



\noindent\textbf{Celebrity}
Once we locate a people/person entity, we check if it's a celebrity. If it is, then who.

\noindent\textit{Check Celebrity}
 \begin{lstlisting}[backgroundcolor = \color{backcolor}, basicstyle=\small, breaklines=true, numbers=none]
"""
USER: <image>
Are you confident that this person is famous?
Answer with yes or no: ASSISTANT: 
"""
\end{lstlisting}

\noindent\textit{Get Celebrity Name}
 \begin{lstlisting}[backgroundcolor = \color{backcolor}, basicstyle=\small, breaklines=true, numbers=none]
"""
USER: <image>
What is the name of this celebrity that you are most confident with? ASSISTANT: 
"""
\end{lstlisting}


\subsection{Reasoner Prompts}




\noindent\textit{Concrete Keywords}
 \begin{lstlisting}[backgroundcolor = \color{backcolor}, basicstyle=\small, breaklines=true, numbers=none]
"""
"{event}"
"{keywords}"
Above texts are describing an same event. Please conclude the event with no more than 5 words. Keep your answer informative, short and concise. 
Short Answer:
"""
\end{lstlisting}

\noindent\textit{Get Location Candidate}
 \begin{lstlisting}[backgroundcolor = \color{backcolor}, basicstyle=\small, breaklines=true, numbers=none]
"""
"{event}"
{keywords}"
Do any of the above text contains location information such as city, country or continent? If yes, please answer that city, country or continent. Otherwise, answer No. 
Short Answer:
"""
\end{lstlisting}




\noindent\textit{Get Time Candidate}
 \begin{lstlisting}[backgroundcolor = \color{backcolor}, basicstyle=\small, breaklines=true, numbers=none]
"""
"{time_clue}"
Given above information, please ground a specific year or month or date from it and answer with it only. If you can not find a date, answer No. 
Short Answer:
"""
\end{lstlisting}


\subsection{Noise Filter}

\noindent\textit{Check Confidence of Location}
 \begin{lstlisting}[backgroundcolor = \color{backcolor}, basicstyle=\small, breaklines=true, numbers=none]
"""
USER: <image>
Are you confident this image was taken in {loc}? 
Short Answer: ASSISTANT: 
"""
\end{lstlisting}

\noindent\textit{Check Confidence of Date}
 \begin{lstlisting}[backgroundcolor = \color{backcolor}, basicstyle=\small, breaklines=true, numbers=none]
"""
USER: <image>
Are you confident this image was taken on {date}? 
Short Answer: ASSISTANT: 
"""
\end{lstlisting}



\begin{table*}[t]
  \centering
  
  \begin{tabular}{lcccc}
    \toprule
         &    \multicolumn{2}{c}{\textbf{Time}}
         &\multicolumn{2}{c}{\textbf{Location}} \\
    \midrule
    \textbf{Model}  &\textbf{Acc(\%)}            & \textbf{X-F1}$^\beta$      & \textbf{Std. Acc (\%)} & \textbf{X-F1}$^\beta$ \\
    BLiP2        &  0.30   & 32.27                & \textit{17.41} & \textit{43.59}                 \\
    LLaVA        &  0.23   & 43.26       & 7.85           & 25.92    \\
    \midrule
    PuzzleGPT       & 0.30   & \textbf{43.72} & \textbf{22.99 }  &  \textbf{51.04}  \\
    PuzzleGPT(LLaVA)       & 0.30   & \textit{43.46} & 13.71   &  31.92  \\
    \bottomrule
    
  \end{tabular}
  \caption{\label{tab:withllava}
Performance drops when switching from BLiP2 to LLaVA in PuzzleGPT. We discovered a significant drop on location performance, which is consistent to the location performance gap between BLiP2 and LLaVA. With a stronger perceiver, a better performance might be expected.}
\end{table*}
\section{Open-Source LLama as the Reasoner}
\label{sec:llama_result}
\begin{table}[t]
  \centering
  \captionsetup{skip=3pt}
\fontsize{9}{11}\selectfont
\setlength{\tabcolsep}{4pt}
  \begin{tabular}{l|cccc}
    \toprule
    &    \multicolumn{2}{c}{\textbf{Time}}&\multicolumn{2}{c}{\textbf{Location}}\\
    \textbf{Reasoner}  &    \textbf{ Acc(\%)}    & \textbf{X-F1$^\beta$} &\textbf{Std. Acc(\%)} & \textbf{X-F1$^\beta$}  \\
    \midrule
    ChatGPT  &\textbf{0.30}  &\textbf{43.72}  & \textbf{22.99}  &\textbf{56.11}   \\
    Llama 3.1 8B &0.15& 27.09& 7.53 & 20.18\\ 
    \bottomrule
  \end{tabular}
  \caption{\label{tab:reasoner_ablation}
   We ablate the models we use for the reasoner. The result shows our method with ChatGPT as reasoner significantly outperforms Llama 3.1.}
   \vspace{-0.3cm}
\end{table}

We seek to understand what would happen if we used open-sourced LLMs as the Reasoner instead of proprietary GPT models.
As such, we compare a state-of-the-art LLaMA 3.1 \cite{touvron2023llamaopenefficientfoundation} against ChatGPT. We couldn't compare it against the larger LLaMA version due to limited compute availability. As shown in Table \ref{tab:reasoner_ablation}, Llama 3.1 performs poorly against ChatGPT. ChatGPT as a reasoner achieves 0.30 and 22.99 in Time and Location (standardized) accuracy, and X-F1\textsuperscript{$\beta$} of 43.72 and 56.11. ChatGPT significantly outperforms the result with Llama 3.1 in both tasks and metrics, indicating open-source models may not be best suited for our task.

\section{Perceiver Ablation}
\label{sec:perceiver_ablation}
We ablated the performance of our BLIP-2 perceiver by replacing it with LLaVA. The results are shown in \Cref{tab:withllava}. Using BLiP2 as the perceiver outperformed using LLaVA, especially on location scores. This might be due to LLaVA's worse performance on location reasoning compared to BLiP2. For time Example-F1, using LLaVA as perceiver scored 43.46 with a brevity penalty, which is still better than the LLaVA baseline. This suggests using different backbones as the perceiver will inherently affect the models' output nature but generally elevate the performance compared to the backbone.

\section{Experiment on a Small Clean Subset}
\label{sec:experiment_small}
One potential reason for the VLMs' consistently low performance on the TARA dataset could be its inherent difficulty, even for humans, in inferring the time and location from the images. To explore this, we manually selected 50 data points where the images were considered informative and indicative of time and location. We then conducted experiments on this subset. The results, shown in Table \Cref{tab:tara_manual_set}, demonstrate a significant performance improvement for our method, while BLIP2 and LLaVA did not show similar improvements. This suggests that although some data points in TARA are extremely challenging, the consistent marginal performance of BLIP2 and LLaVA indicates their inability to effectively handle the dataset's visual clues. In contrast, PuzzleGPT exhibited a notable improvement, highlighting its robustness and superior ability to utilize information from the image.
\section{Qualitative Analysis}
\label{sec:qualitative_analysis}
\begin{figure*}[t]
\centering
  \includegraphics[width=\linewidth]{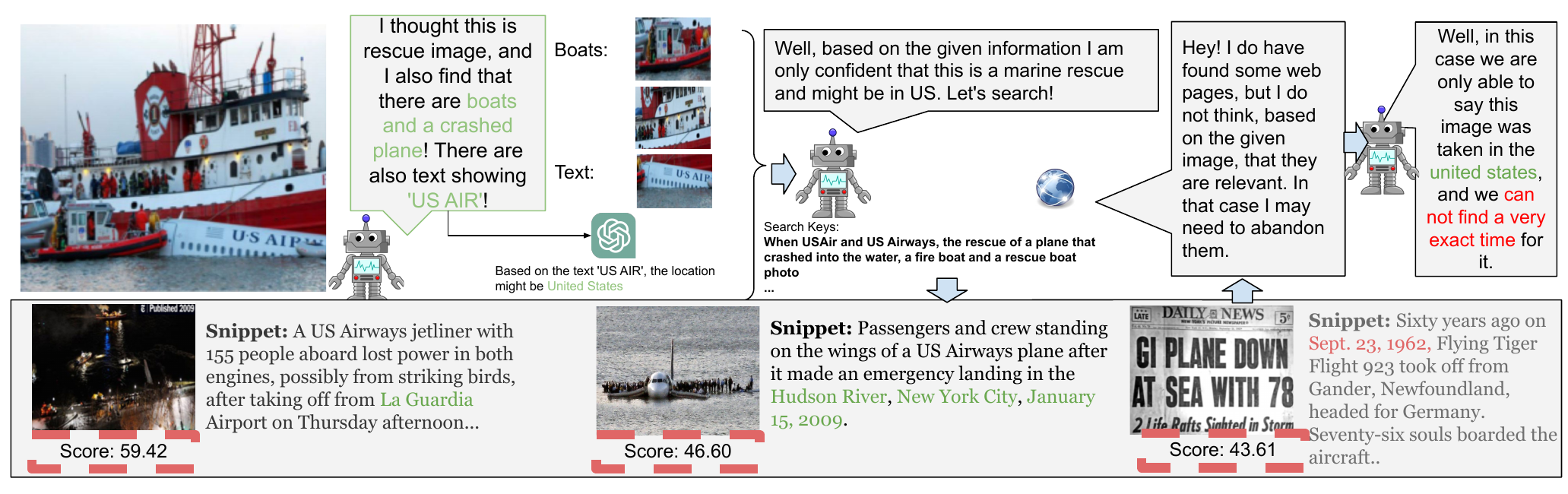}\\
  \vspace{2.5em}
  \includegraphics[width=\linewidth]{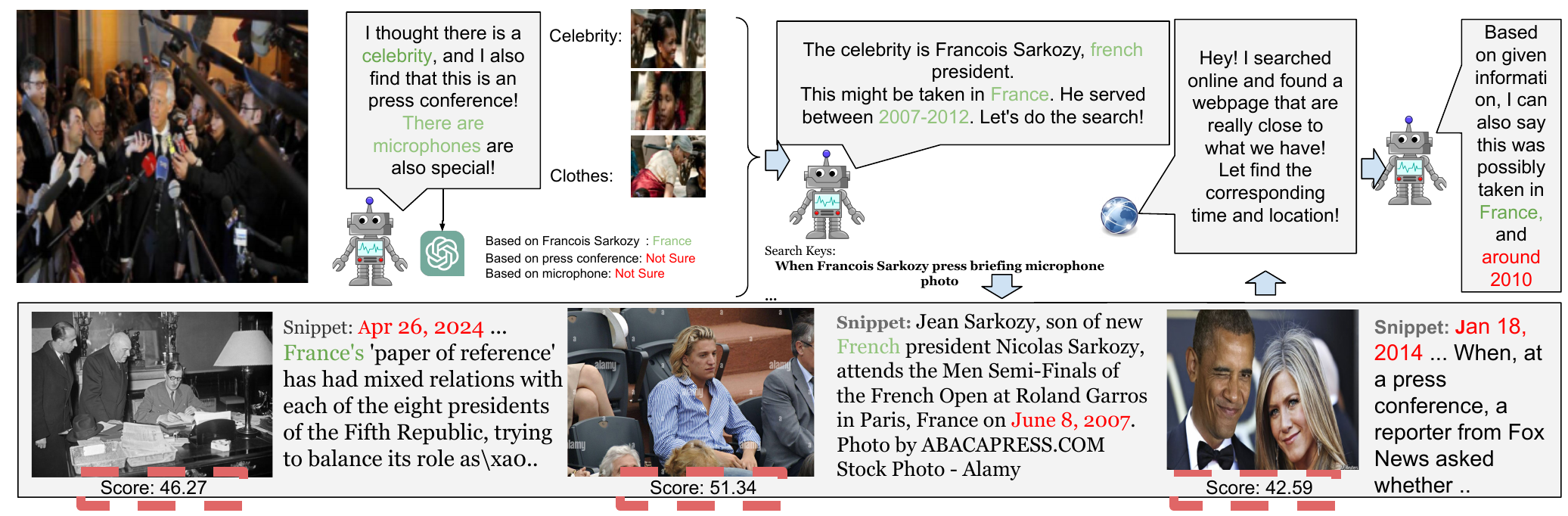}

  \caption {Two samples for negative case studies.}
  \label{fig:neg}
\end{figure*}

\begin{figure*}[t]
\centering
  \includegraphics[width=\linewidth]{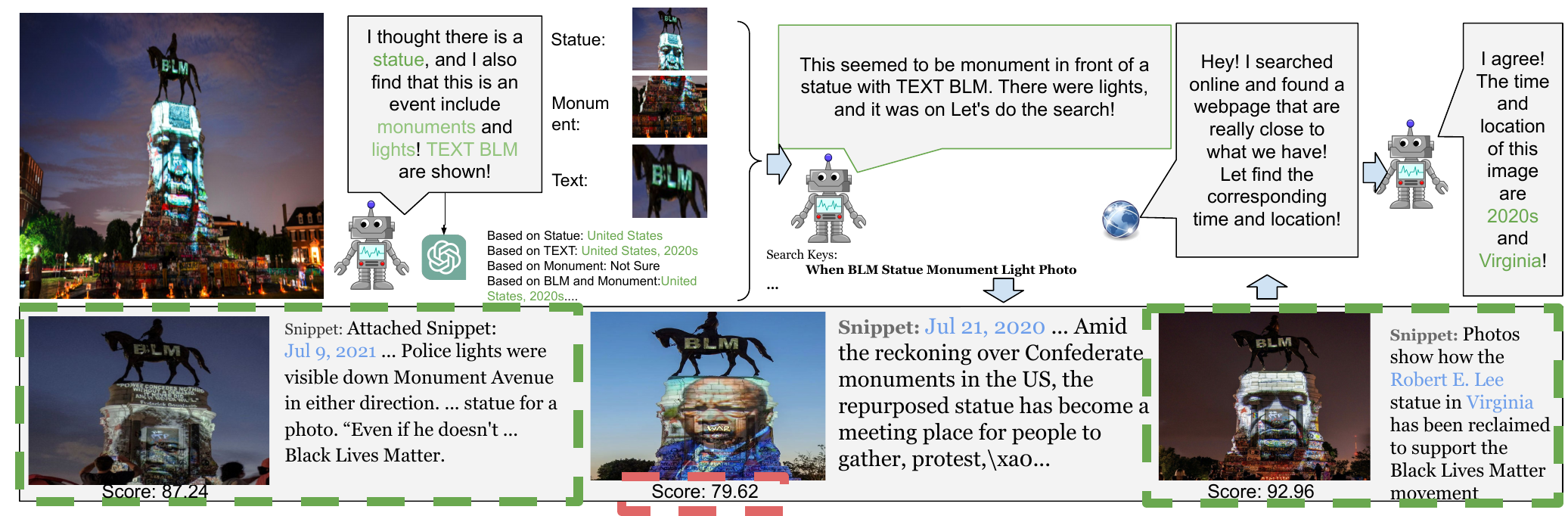}
  \caption {Another positive case study.}
  \label{fig:ps2}
\end{figure*}

We conducted case studies on TARA for qualitative analysis to further demonstrate PuzzleGPT's effectiveness. In Figures \ref{fig:neg}, and \ref{fig:ps2}, we show some additional positive and negative case studies of our system.
Our analysis of failure cases led us to observe that failure typically occurs when the given image lacks unique landmarks, events, or people. For example, a generic image showing a crashed plane (\Cref{fig:neg} top) gives little clue about the specific location. Or a politician speaking at a press conference (\Cref{fig:neg} bottom) provides little to no evidence to deduce time.
\begin{table}[t]
\fontsize{9}{11}\selectfont
  \centering
  
  \begin{tabular}{lcccc}
    \toprule
         &    \multicolumn{2}{c}{\textbf{Time}}
         &\multicolumn{2}{c}{\textbf{Location}} \\
    \midrule
    \textbf{Model}  &\textbf{Acc(\%)}            & \textbf{X-F1}$^\beta$      & \textbf{Acc(\%)} & \textbf{X-F1}$^\beta$ \\
    \midrule
    BLiP2      & 4.88    & 38.59     & \textit{22.92}  & 38.55  \\
    LLaVA      & \textit{9.76}    & \textit{43.25} & \textit{22.92}    & \textit{40.34}   \\
    \midrule
    PuzzleGPT       & \textbf{12.20}   & \textbf{46.24} & \textbf{35.42}   &  \textbf{57.75}  \\
    \bottomrule
    
  \end{tabular}
  \caption{\label{tab:tara_manual_set}
Experiments conducted on a \textbf{smaller (50)} subset that are manually selected by the human evaluator. Instances in this dataset are considered information-rich, while generative VLMs failed to receive performance improvement.
  }
\end{table}
\begin{table*}[t]
  \centering
    \captionsetup{skip=3pt}
  \setlength{\tabcolsep}{4pt}
  \fontsize{9}{11}\selectfont
  \begin{tabular}{lccccccccc}
    \toprule
         &   \multicolumn{3}{c}{\textbf{Time}}&\multicolumn{3}{c}{\textbf{Country}}&\multicolumn{3}{c}{\textbf{Region}} \\
    \midrule
    \textbf{Models}  &\textbf{Acc(\%)}     & \textbf{Prec}     & \textbf{F1}     & \textbf{Acc(\%)} & \textbf{Prec}     & \textbf{F1} & \textbf{Acc(\%)} & \textbf{Prec}&\textbf{F1}\\
    \midrule
    OpenFlamingo Test   & 27.70   &  26.36 & 11.49   & 3.89  & 3.69   & 2.18 
                              & 4.72    &  8.62  & 4.72  \\
    OpenFlamingo-VQA          & 31.59   &  30.36 & 28.60   & \textbf{48.88} & \textit{53.78} &\textit{ 41.19}                           &22.49 & 30.49 & 18.64\\
    OpenFlamingo-VQA CoT      & 35.21   &  29.36 & 28.42   & 40.3  & 45.24 & 33.17                           & \textit{24.04} & \textit{39.39} & \textit{19.27}\\
    LLaMA-AdapterV2-Instr$^a$   &\textit{ 58.02}     &  28.04 & 32.88   & 23.05 & 52.64 & 18.66                           & 19.07 & 26.59 & 13.01\\
    LLaMA-AdapterV2-Instr$^b$   & 34.34     &  \textit{58.59} & \textit{37.70}   & \textit{45.62} & 51.57 & 35.50                           & 11.12 & 10.05 & 5.99\\
    BLIP2 &51.11&54.84&42.51&50.95&30.48&24.97&51.26&51.68&42.88\\
    InstructBLIP &38.89&2.71&1.76&27.62&25.47&12.52&44.44&24.69&19.43\\
    LLaVA&63.65&27.79&21.90&41.43&30.60&18.04&\textbf{62.22} &43.26&41.15\\
        
    ViperGPT  & 29.52   &  35.27 & 21.50   & 20.32  & 21.96    & 14.98   & 26.03 & 25.88                           & 21.54\\
    
    IdealGPT  & 32.16   &  29.20 & 26.30   & 4.32  &   4.10  &  2.21  & 9.13 & 9.85& 7.68\\
    Frequency Baseline  & 25.07   &  25.29 & 23.27   & 3.33  & 2.95    & 2.88   & 12.53 & 12.59                           & 12.25\\
    \midrule
    PuzzleGPT(Ours)             & \textbf{71.90}   &  \textbf{70.63}     & \textbf{72.61}       & 43.65  & \textbf{72.78}  & \textbf{49.79} & 62.06 &    \textbf{79.22}         & \textbf{68.18}\\
    \bottomrule
  \end{tabular}
  \caption{\label{tab:wikitilo_more}
PuzzleGPT generalizes to WikiTilo dataset, scoring state-of-the-art performance in almost all the metrics.
  }
\end{table*}

\section{Additional WikiTilo Baselines}
\label{sec:additional_wikitilo_baselines}

We compare PuzzleGPT to additional Multimodal LLMs-based baselines in \Cref{tab:wikitilo_more}. We find that PuzzleGPT consistently performs better than these additional baselines as well on all categories. Only LLaVA performs marginally better on Region Acc by 0.2\%. Still, the F1 Region score for PuzzleGPT is \textasciitilde 15\% better than LLaVA.


\section{Ablation Metrics}
\label{sec:metrics}
The accuracy metric measures the exact matching of open-ended labels in TARA. As such, all models record lower accuracy as compared to the more relaxed F1-based metric (as illustrated in \Cref{tab:tara_main_results_zeroshot}, \Cref{tab:tara_main_results_finetune} and the original TARA dataset paper). The minor difference in accuracy metric makes ablations tricky and hard to assess. The strengths/weaknesses of our components are better reflected in the F1-based metric. As such, we report F1-based metrics for all our ablations.

\end{document}